\def\year{2018}\relax
\documentclass[letterpaper]{article} %DO NOT CHANGE THIS
\usepackage{aaai18}  %Required
\usepackage{times}  %Required
\usepackage{helvet}  %Required
\usepackage{courier}  %Required
\usepackage{url}  %Required
\usepackage{graphicx}  %Required
\usepackage{subfigure}
\usepackage{amsmath}
\usepackage{amsfonts}
\usepackage{amssymb}

\begin{document}
\title{Variational Inference: A Unified Framework of \\ Generative Models and Some Revelations}
\author{Jianlin Su\\
School of Mathematics, Sun Yat-sen University
}
\maketitle
\begin{abstract}
We reinterpreting the variational inference in a new perspective. Via this way, we can easily prove that EM algorithm, VAE, GAN, AAE, ALI(BiGAN) are all special cases of variational inference. The proof also reveals the loss of standard GAN is incomplete and it explains why we need to train GAN cautiously. From that, we find out a regularization term to improve stability of GAN training.
\end{abstract}

\noindent In recent years, deep generative models, espcially Generative Adversarial
Networks (GANs) \cite{Goodfellow2014Generative}, have achieved impressive success. We can find dozens of different variants of GANs. However, most of them are achieved empirically, rarely have completely theoretical guidance.

This paper aims to establish a unified framework of these various generative models by variational inference. Firstly, we present a new formulation of variational inference, which can derive EM algorithm and Variational Autoencoders (VAEs) \cite{Kingma2013Auto} in only serveral lines. Then we re-derive GAN by our new variational inference and find the loss of standard GAN is not complete, which lacks of a regularization term. Without this term, we need to adjust hyperparameters carefully to make GAN converge.

In fact, the original purpose of our work is to incorporate GAN into the variational inference framework. It seems we are successful to accomplish it. The new regularization term is an unexpected result. Fortunately, we are glad to see it autually work in our experiment.

\section{Variational Inference}

Suppose $x$ is an explicit variable, $z$ is a latent variable，and $\tilde{p}(x)$ is evidence distribution of $x$. We let
\begin{equation}q_{\theta}(x)=\int q_{\theta}(x,z)dz\end{equation}
and we hope $q_{\theta}(x)$ will be a good approximation of $\tilde{p}(x)$. In general cases, we want to maximize log likelihood function
\begin{equation}\theta = \mathop{\arg\max}_{\theta}\, \int \tilde{p}(x)\log q(x)dx\end{equation}
which is equivalent to minimizing $KL(\tilde{p}(x)\Vert q(x))$:
\begin{equation}KL(\tilde{p}(x)\Vert q(x)) = \int \tilde{p}(x) \log \frac{\tilde{p}(x)}{q(x)}dx\end{equation}
But if we can not calculate the integral analytically, we can not maximize log likelihood or minimize KL-divergence directly.

The variational inference changes objective function: rather minimizing KL-divergence of marginal distributions $KL(\tilde{p}(x)\Vert q(x))$, we can minimize the KL-divergence of joint distribution $KL(p(x,z)\Vert q(x,z))$ or $KL(q(x,z)\Vert p(x,z))$. We have
\begin{equation}\begin{aligned}&KL(p(x,z)\Vert q(x,z))\\
=& KL(p(x)\Vert q(x)) + \int p(x) KL(p(z|x)\Vert q(z|x)) dx\\
\geq& KL(p(x)\Vert q(x)),\end{aligned}\end{equation}
which suggests $KL(p(x,z)\Vert q(x,z))$ is an upper bound of $KL(\tilde{p}(x)\Vert q(x))$. In many cases, joint KL-divergence easier to calculate than marginal KL-divergence. Therefore, variational inference provides a computable solution. If it works, we have $q(x,z)\to p(x,z)$，which means $q(x) = \int q(x,z)dz \to \int p(x,z)dz = \tilde{p}(x)$. Namely, $q(x)$ becomes an approximation of the real distribution $\tilde{p}(x)$.

\section{VAE and EM algorithm}

Due to our new insight of variational inference, VAE and EM algorithm can be derived in a very simple way.

In VAE, we let $q(x,z)=q(x|z)q(z), p(x,z)=\tilde{p}(x) p(z|x)$, while $q(x|z),p(z|x)$ are Gaussian distributions with unknown parameters and $q(z)$ is standard Gaussian distribution. The loss we need to minimize is
\begin{equation}\label{eq:kl-o} \begin{aligned}&KL\left(p(x,z)\Vert q(x,z) \right)\\
=&\iint \tilde{p}(x) p(z|x) \log \frac{\tilde{p}(x) p(z|x)}{q(x|z)q(z)}dxdz\end{aligned}\end{equation}
while $\log \tilde{p}(x)$ does not contain any parameters, it does not change final result. So loss can be transed into
\begin{equation}\mathbb{E}_{x\sim \tilde{p}(x)}\left[-\int p(z|x)\log q(x|z)dz + KL(p(z|x)\Vert q(z))\right]\end{equation}
Because $q(z),p(z|x)$ are both Gaussian, we can get the analytic expression of $KL(p(z|x)\Vert q(z))$. And with the reparametrization trick, the first term can be approximate as $\log q(x|z)$. Consequently, the final loss for VAE is
\begin{equation}\begin{aligned}\mathbb{E}_{x\sim \tilde{p}(x)}\Big[-\log q(x|z) + KL(p(z|x)\Vert q(z))\Big]\end{aligned}\end{equation}

The assumption of EM algorithm is like VAE, excluding supposing $p(z|x)$ is Gaussian. In EM algorithm, the loss is still $\eqref{eq:kl-o}$, but we treat entire $p(z|x)$ as training parameters. Rather than minimizing the loss directly, here we use an alternate training way. Firstly, we fix $p(z|x)$ and just optimize $q(x|z)$. Removing the "constant" term, the loss of $q(x|z)$ is
\begin{equation}\label{eq:em-1}q(x|z) = \mathop{\arg\max}_{q(x|z)} \,\mathbb{E}_{x\sim \tilde{p}(x)}\left[\int p(z|x) \log  q(x,z) dz\right]\end{equation}
Secondly, we fix $q(x|z)$ and optimize $p(z|x)$. We define $q(z|x)$ by $q(x|z)$
\begin{equation}q(x)=\int q(x|z)q(z)dz,\quad q(z|x)=\frac{q(x|z)q(z)}{q(x)}\end{equation}
now we have
\begin{equation}\begin{aligned}p(z|x) =& \mathop{\arg\min}_{p(z|x)} \,\mathbb{E}_{x\sim \tilde{p}(x)}\left[\int p(z|x) \log \frac{p(z|x)}{q(z|x)q(x)} dz\right]\\
=& \mathop{\arg\min}_{p(z|x)} \,\mathbb{E}_{x\sim \tilde{p}(x)}\left[KL\left(p(z|x)\Vert q(z|x)\right)-\log q(x)\right]\\
=& \mathop{\arg\min}_{p(z|x)} \,\mathbb{E}_{x\sim \tilde{p}(x)} \left[KL\left(p(z|x)\Vert q(z|x)\right)\right]
\end{aligned}\end{equation}
Because we don't make any assumptions about the form of $p(z|x)$, we can let $p(z|x) = q(z|x)$ make loss equal zero, which is an optimal solution of $p(z|x)$. In other words, the optimal $p(z|x)$ is
\begin{equation}\label{eq:em-2}p(z|x) = \frac{q(x|z)q(z)}{\int q(x|z)q(z)dz}\end{equation}
EM algorithm is just to perform $\eqref{eq:em-1},\eqref{eq:em-2}$ alternately。

\section{GAN within Variational Inference}

In this section, we describe a general approach to incorporate GAN into the variational inference, which leads a new insight to GAN and results a effective regularization for GAN.

\subsection{General Framework}

As same as VAE, GAN also want to achive a generative model $q(x|z)$, which can transform $z\sim q(z)=N(z;0,I)$ to the evidence distribution $x \sim \tilde{p}(x)$. Different from Gaussian assumption in VAE, GAN let $q(x|z)$ be a Dirac delta function
\begin{equation}q(x|z)=\delta\left(x - G(z)\right),\quad q(x)=\int q(x|z)q(z)dz\end{equation}
whose $G(z)$ is a neutral network of generative model, called generator.

Generally, we considered $z$ is a random latent variable in generative model. However, it is well-known that Dirac delta function is non-zero at only one point, so the mapping from $z$ to $x$ in GAN is almost one to one. The variable $z$ is not "random" enough, so we do not treat it as a latent variable (that means we need not to consider posterior distribution $p(z|x)$). In fact, we just consider the binary random variable $y$ as a random latent variable in GAN:
\begin{equation}q(x,y)=\left\{\begin{aligned}&\tilde{p}(x)p_1,\,y=1\\&q(x)p_0,\,y=0\end{aligned}\right.\end{equation}
here $p_1 = 1-p_0$ discribing a Bernoulli distribution. For simpler we set $p_1=p_0=1/2$. 

On the other hand, we let $p(x,y)=p(y|x) \tilde{p}(x)$, while $p(y|x)$ is a conditional Bernoulli distribution. Distinct from VAE, GAN choose another direction of KL-divergence as optimal objective:
\begin{equation}\begin{aligned}&KL\left(q(x,y)\Vert p(x,y) \right)\\
=&\int \left(\tilde{p}(x)p_1\log \frac{\tilde{p}(x)p_1}{p(1|x)\tilde{p}(x)}+ q(x)p_0\log \frac{q(x)p_0}{p(0|x)\tilde{p}(x)}\right)dx\\
\sim&\int \tilde{p}(x)\log \frac{1}{p(1|x)}dx+\int q(x)\log \frac{q(x)}{p(0|x)\tilde{p}(x)}dx\end{aligned}\end{equation}
Once succeed, we have $q(x,y)\to p(x,y)$, means
\begin{equation}\begin{aligned}&p_1 \tilde{p}(x) + p_0 q(x) \\
=& \sum_y q(x,y) \to \sum_y p(x,y) = \tilde{p}(x)\end{aligned}\end{equation}
consequently $q(x)\to\tilde{p}(x)$.

Now we have to solve $p(y|x)$ and $G(x)$. For simpler we set $p(1|x)=D(x)$, called discriminator. Like EM algorithm, we use a alternately training strategy. Firstly, we fix $G(z)$, so $q(x)$ does. Ignoring constants for $G(z)$, we get:
\begin{equation}\begin{aligned}D = \mathop{\arg\min}_{D} &-\mathbb{E}_{x\sim\tilde{p}(x)}\left[\log D(x)\right]\\
&-\mathbb{E}_{x\sim q(x)}\left[\log (1-D(x))\right]\end{aligned}\end{equation}
Then we fix $D(x)$ for optimizing $G(x)$. Ignoring constants for $D(x)$, we get the pure loss:
\begin{equation}\label{eq:gan-g-loss}G = \mathop{\arg\min}_{G}\int q(x)\log \frac{q(x)}{(1-D(x)) \tilde{p}(x)}dx\end{equation}
For minimizing this loss, we need the formula of $\tilde{p}(x)$, which is always impossible. For the same reason as $\eqref{eq:em-2}$, if $D(x)$ has enough fitting ability, the optimal $D(x)$ is
\begin{equation}D(x)=\frac{\tilde{p}(x)}{\tilde{p}(x)+q^{o}(x)}\end{equation}
$q^{o}(x)$ is $q(x)$ at previous stage. We can solve $\tilde{p}(x)$ from it and replace $\tilde{p}(x)$ in $\eqref{eq:gan-g-loss}$:
\begin{equation}\label{eq:g-loss}\begin{aligned}&\int q(x)\log \frac{q(x)}{D(x)  q^{o}(x)}dx\\
=&-\mathbb{E}_{x\sim q(x)}[\log D(x)] + KL\left(q(x)\Vert  q^{o}(x)\right)\\
=&-\mathbb{E}_{z\sim q(z)}[\log D(G(z))] + KL\left(q(x)\Vert  q^{o}(x)\right)
\end{aligned}\end{equation}

\subsection{Basic Analysis}

It is obviously that the fisrt term is one of the standard $G$ losses of GAN:
\begin{equation}-\mathbb{E}_{z\sim q(z)}\log D(G(z))\end{equation}
The second extra item describes the distance between the new distribution and the old distribution. Two terms are adversarial. $KL\left(q(x)\Vert  q^{o}(x)\right)$ try to make the two distributions more similar, while $-\log D(x)$ will be very large because $D(x)$ will be very small for $x\sim q^{o}(x)$ if discriminator is trained fully (all of them will be considered as negative samples), and vice versa. Thus, minimizing entire loss requires model to inherit the old distribution $q^{o}(x)$ and explore the new world $p(1|y)$.

As we know, the generator's loss in current standard GAN has no the second term, which is autually an incomplete loss. Suppose there is a omnipotent optimizer which can identify the global optimum in very short time and $G(z)$ has enough fitting ability, then $G(z)$ can only generate just one sample which make $D(x)$ largest. In other words, the global optimal solution of $G(z)$ is $G(z)=x_0$, while $x_0=\mathop{\arg\max}_{x} D(x)$. That is called Model Collapse, which will occur certainly in theory.

So, what enlightenment can $KL\left(q(x)\Vert q^{o}(x)\right)$ give for us? We let
\begin{equation}q^{o}(x)=q_{\theta-\Delta \theta}(x),\quad q(x)=q_{\theta}(x)\end{equation}
that means the updates of parameters of $G(z)$ in this iteration is $\Delta\theta$. Using Taylor series to expand $q_{\theta-\Delta \theta}(x)$ to second order, we get
\begin{equation}\begin{aligned}KL\left(q(x)\Vert q^{o}(x)\right)&\approx \int\frac{\left(\Delta\theta\cdot \nabla_{\theta}q_{\theta}(x)\right)^2}{2q_{\theta}(x)} dx \\
&\approx \left(\Delta\theta\cdot c\right)^2\end{aligned}\end{equation}
We have already indicated that a complete loss should contain $KL\left(q(x)\Vert q^{o}(x)\right)$. If not, we should keep it small during training by other way. The above approximation shows the extra loss is about $\left(\Delta\theta\cdot c\right)^2$, which can not be too large, meaning $\Delta\theta$ can not be too large because $c$ can be regard as a constant during one iteration. 

Now we can explain why we need to adjust hyperparameters carefully to make GAN converge\cite{Salimans2016Improved}. The most common optimizers we use are all based on gradient descent, so $\Delta\theta$ is proportional to gradients. We need to keep $\Delta\theta$ small, which is equivalent to keep gradients small. Consequently, we apply gradients clipping, Batch Normalization in GAN because they can make gradients steady. For the same reason, we always use Adam rather than SGD+Momentum. Meanwhile, the iterations of $G(z)$ can not be too many, while $\Delta\theta$ will be large if $G(z)$ updates a lot.

\subsection{Regularization Term}

Here we focus on getting something really useful and practical. We try to calculate $KL\left(q(x)\Vert  q^{o}(x)\right)$ directly, for obtain an regularization term to add in generator's loss. Because of difficulty of directly calculating, we estimate it via calculating $KL\left(q(x,z)\Vert \tilde{q}(x,z)\right)$ (maybe inspired by variational inference):
\begin{equation}\begin{aligned}&KL\left(q(x,z)\Vert \tilde{q}(x,z)\right)\\
=&\iint q(x|z)q(z)\log \frac{q(x|z)q(z)}{\tilde{q}(x|z)q(z)}dxdz\\
=&\iint \delta\left(x-G(z)\right)q(z)\log \frac{\delta\left(x-G(z)\right)}{\delta\left(x-G^{o}(z)\right)}dxdz\\
=&\int q(z)\log \frac{\delta(0)}{\delta\left(G(z)-G^{o}(z)\right)}dz
\end{aligned}\end{equation}
we have a limitation
\begin{equation}\delta(x)=\lim_{\sigma\to 0}\frac{1}{(2\pi\sigma^2)^{d/2}}\exp\left(-\frac{x^2}{2\sigma^2}\right)\end{equation}
which means $\delta(x)$ can be replaced with a Gaussian distribution of small variance. So we have
\begin{equation}KL\left(q(x)\Vert  q^{o}(x)\right)\sim \lambda \int q(z)\Vert G(z) - G^{o}(z)\Vert^2 dz\end{equation}
$\eqref{eq:g-loss}$ becomes
\begin{equation}\mathbb{E}_{z\sim q(z)}\left[-\log D(G(z))+\lambda \Vert G(z) - G^{o}(z)\Vert^2\right] \end{equation}
In other words, we can use the distance between samples from old and new generator as a regularization term, to guarantee the new generator has little deviation from old generator.

\begin{figure*}
  \centering
  \subfigure[With our regularization, the model keeps steady.]{
    \label{fig:face_normal_me}
    \includegraphics[width=16cm]{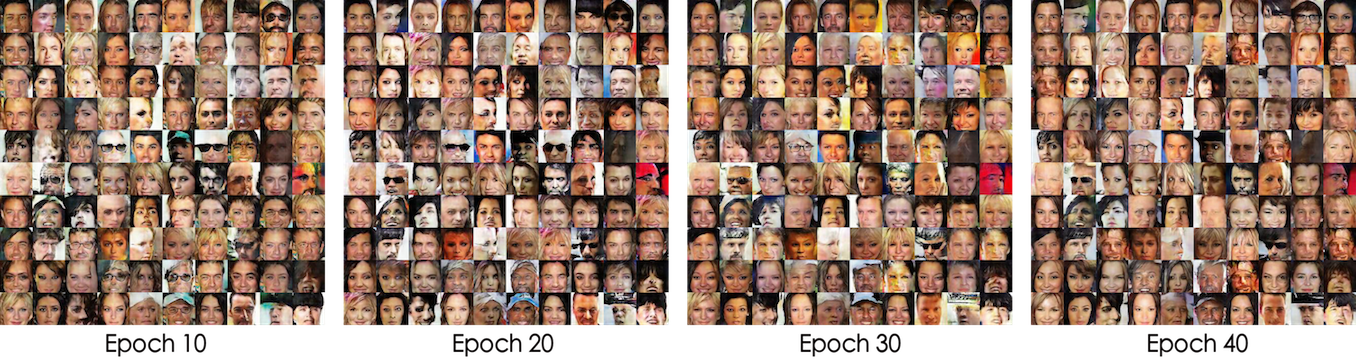}}
  \hspace{1in}
  \subfigure[Without regularization, the model collapses after 25 epochs.]{
    \label{fig:face_normal_old}
    \includegraphics[width=16cm]{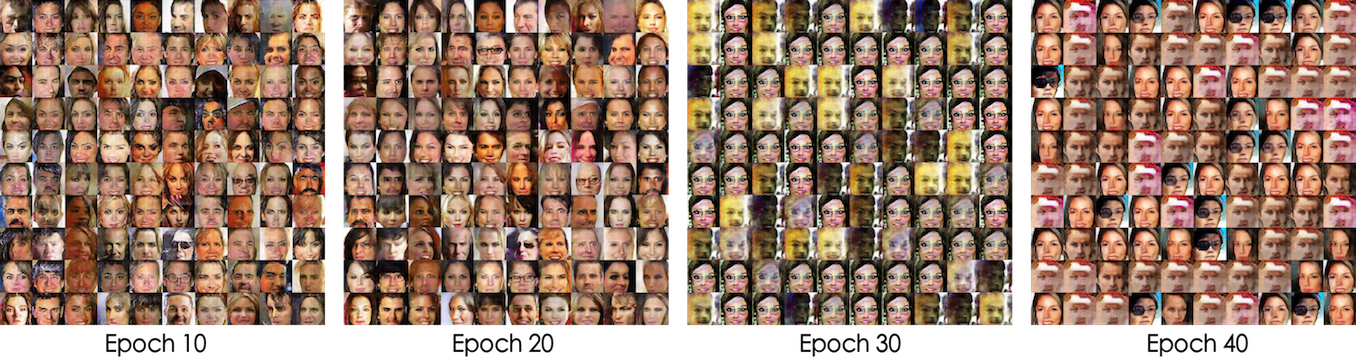}}
  \caption{An ordinary DCGAN model on CelebA, train discriminator and generator one iteration per period.}
  \label{fig:face_normal}
\end{figure*}

\begin{figure*}
  \centering
  \subfigure[With our regularization, the model has a faster convergence.]{
    \label{fig:face_nobn_me}
    \includegraphics[width=16cm]{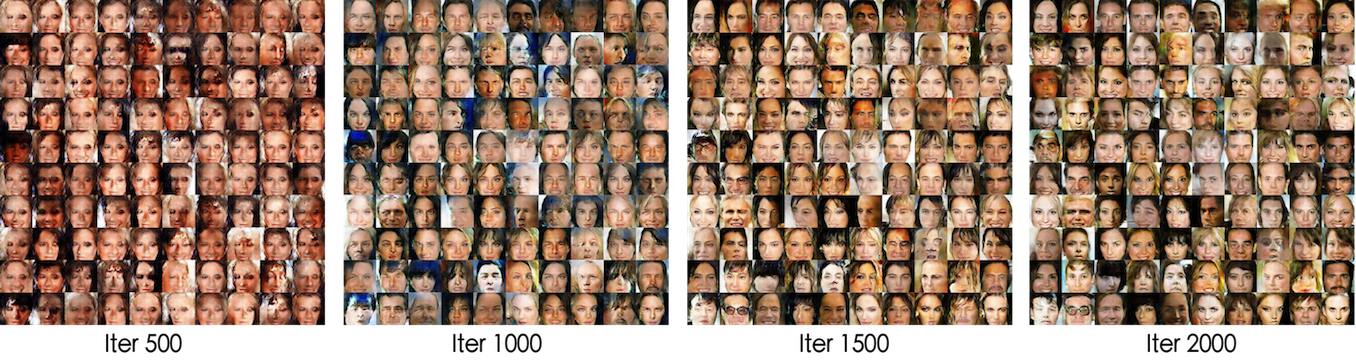}}
  \hspace{1in}
  \subfigure[Without regularization, the model need more iterations for convergence.]{
    \label{fig:face_nobn_old}
    \includegraphics[width=16cm]{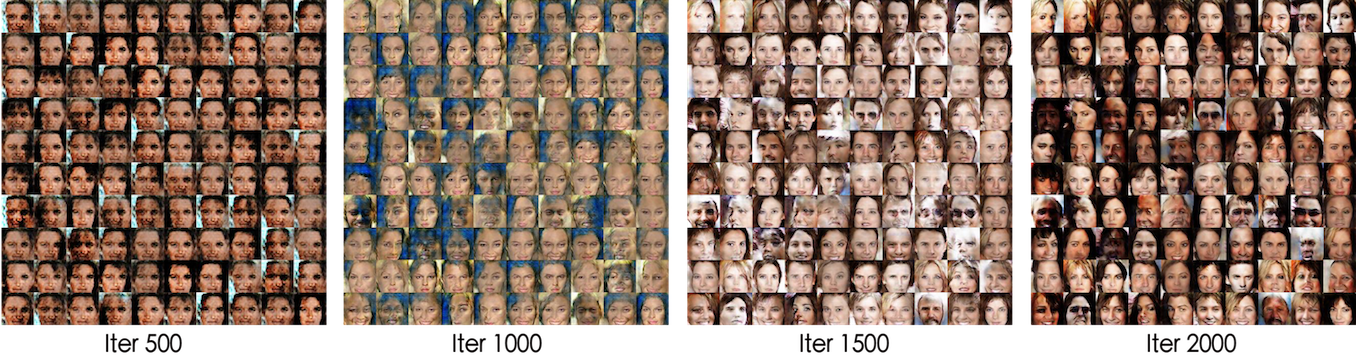}}
  \caption{An ordinary DCGAN model removing BN, train discriminator and generator 5 iterations per period.}
  \label{fig:face_nobn}
\end{figure*}

Experiment $\ref{fig:face_normal}$ and $\ref{fig:face_nobn}$ on CelebA datasets\footnote{the code is modified from \url{https://github.com/LynnHo/DCGAN-LSGAN-WGAN-WGAN-GP-Tensorflow}, now available at \url{https://github.com/bojone/gan/tree/master/vgan}.} shows this regularization works well.

\section{Models Related to GAN}

Adversarial Autoencoders (AAE)\cite{Makhzani2015Adversarial} and Adversarially Learned Inference (ALI)\cite{Dumoulin2016Adversarially} are two variants of GAN, they also can be incorporated into variational inference. Of course, with the preparation above, it is just like two homework questions.

\subsection{AAE under GAN framework}

Autually, for obtaining AAE, the only thing we need to do is exchanging $x,z$ in standard GAN. In detail, AAE wants to train an encoder $p(z|x)$ to map the distribution of real data $\tilde{q}(x)$ to the standard Gaussian distribution $q(z)=N(z;0,I)$, while
\begin{equation}p(z|x)=\delta\left(z - E(x)\right),\quad p(z)=\int p(z|x)\tilde{q}(x)dx\end{equation}
whose $E(x)$ is a neutral network of encoder.

Like GAN, AAE needs a binary random latent variable $y$, and
\begin{equation}p(z,y)=\left\{\begin{aligned}&p(z)p_1,\,y=1\\&q(z)p_0,\,y=0\end{aligned}\right.\end{equation}
we also let $p_1=p_0=1/2$. On the other hand, we set $q(z,y)=q(y|z) q(z)$, whose posterior distribution $p(y|z)$ is conditional Bernoulli distribution taking $z$ as input. Now we minimize $KL\left(p(z,y)\Vert q(z,y) \right)$:
\begin{equation}\begin{aligned}&KL\left(p(z,y)\Vert q(z,y) \right)\\
=&\int p(z)p_1\log \frac{p(z)p_1}{q(1|z)q(z)}dz+\int q(z)p_0\log \frac{q(z)p_0}{q(0|z)q(z)}dz\\
\sim&\int p(z)\log \frac{p(z)}{q(1|z)q(z)}dz+\int q(z)\log \frac{1}{q(0|z)}dz\end{aligned}\end{equation}
Now we have to solve $q(y|z)$ and $E(x)$. we set $p(0|z)=D(z)$ and still train it alternately. Firstly, we fix $E(x)$, so $p(z)$ does. Ignoring constants for $E(z)$, we get:
\begin{equation}\begin{aligned}D=\mathop{\arg\min}_D &-\mathbb{E}_{z\sim p(z)}\left[\log (1-D(z))\right]\\
&-\mathbb{E}_{z\sim q(z)}\left[\log D(z)\right]\end{aligned}\end{equation}
Then we fix $D(z)$ for optimizing $E(z)$. Ignoring constants for $D(z)$, we get the pure loss:
\begin{equation}E = \mathop{\arg\min}_E \int p(z)\log \frac{p(z) }{(1-D(z)) q(z)}dz\end{equation}
Use the theoretical solution $D(z)=q(z)/[p(^{o}z)+q(z)]$ and replace $q(z)$:
\begin{equation}\mathbb{E}_{x\sim \tilde{p}(x)}[-\log D(E(x))] + KL\left(p(z)\Vert  p^{o}(z)\right)\end{equation}

On the one hand, like standard GAN, if we train carefully, we may remove the second term and have
\begin{equation}\mathbb{E}_{x\sim \tilde{p}(x)}[-\log D(E(x))]\end{equation}
on the other hand, we can train a decoder $G(z)$ after finishing the adversarial trainning. However, if our $E(x), G(z)$ has strong enough modeling ability, then we can add a reconstruction error into encoder's loss, which will not interfere the original adversarial optimition of encoder. Therefore, we get a joint loss:
\begin{equation}\begin{aligned}G,E = \mathop{\arg\min}_{G,E}\mathbb{E}_{x\sim \tilde{p}(x)}[&-\log D(E(x))\\
&+\lambda\Vert x - G(E(x))\Vert^2]\end{aligned}\end{equation}

\subsection{Our Version of ALI}

ALI is like a fusion of GAN and AAE. And there is an almost identical version called Bidirectional GAN (BiGAN)\cite{Donahue2017Adversarial}. Compared with GAN, they treats $z$ as a latent variable, so it needs a posterior distribution $p(z|x)$. Concretely, in ALI we have
\begin{equation}q(x,z,y)=\left\{\begin{aligned}&p(z|x)\tilde{p}(x) p_1,\,y=1\\&q(x|z)q(z)p_0,\,y=0\end{aligned}\right.\end{equation}
and $p(x,z,y)=p(y|x,z) p(z|x) \tilde{p}(x)$, then we minimize $KL\left(q(x,z,y)\Vert p(x,z,y) \right)$:
\begin{equation}\begin{aligned}&\iint p(z|x)\tilde{p}(x) p_1\log \frac{p(z|x)\tilde{p}(x) p_1}{p(1|x,z) p(z|x) \tilde{p}(x)}dxdz\\
&+\iint q(x|z)q(z)p_0\log \frac{q(x|z)q(z)p_0}{p(0|x,z) p(z|x) \tilde{p}(x)}dxdz\end{aligned}\end{equation}
which is equivalent to minimize
\begin{equation}\label{eq: ori-loss-ali}\begin{aligned}&\iint p(z|x)\tilde{p}(x)\log \frac{1}{p(1|x,z)}dxdz\\
&+\iint q(x|z)q(z)\log \frac{q(x|z)q(z)}{p(0|x,z) p(z|x) \tilde{p}(x)}dxdz\end{aligned}\end{equation}

Now we have to solve $p(y|x,z),p(z|x),q(x|z)$. we set $p(1|x,z)=D(x,z)$, while $p(z|x)$ is a Gaussian distribution including an encoder $E(x)$ and $q(x|z)$ is an another Gaussian distribution including an generator $G(z)$. Still alternately train it. Firstly we fix $E,G$, the loss related to $D$ is
\begin{equation}\begin{aligned}D=\mathop{\arg\min}_D &-\mathbb{E}_{x\sim\tilde{p}(x),z\sim p(z|x)} \log D(x,z) \\
&- \mathbb{E}_{z\sim q(z),x\sim q(x|z)} \log (1-D(x,z))\end{aligned}\end{equation}
As same as VAE,  the expectation of $p(z|x)$ and $q(x|z)$ can be done using the the reparametrization trick. Now fix $D$ for optimizing $G,E$, and because of cooccurrence of $E,G$, loss $\eqref{eq: ori-loss-ali}$ can not be simplified. But using the theoretical solution of $D$
\begin{equation}D(x,z)=\frac{p^{o}(z|x)\tilde{p}(x)}{p^{o}(z|x)\tilde{p}(x)+q^{o}(x|z)q(z)}\end{equation}
can transform it to
\begin{equation}\begin{aligned}&-\iint p(z|x)\tilde{p}(x)\log D(x, z) dxdz \\
&-\iint q(x|z) q(z)\log D(x, z) dxdz\\
&+\int q(z) KL(q(x|z)\Vert q^o(x|z)) dz \\
&+ \iint q(x|z) q(z)\log \frac{p^o(z|x)}{p(z|x)}dxdz\end{aligned}\end{equation}
Due to Gaussianity of $q(x|z)$ and $p(x|z)$, we can calculate last two term analytically, or ignore them while optimizing it carefully, leading
\begin{equation}\label{eq:our-ali-g}\begin{aligned}&-\iint p(z|x)\tilde{p}(x)\log D(x, z) dxdz \\&-\iint q(x|z) q(z)\log D(x, z) dxdz\end{aligned}\end{equation}
That is our version of ALI, which has little different from the standard ALI. The current popular view is to treat ALI (includes GAN) as a min-max problem. From that, the loss of encoder and generator is
\begin{equation}\label{eq:our-ali-g-o1}\begin{aligned}&\iint p(z|x)\tilde{p}(x)\log D(x, z) dxdz \\
+ &\iint q(x|z) q(z)\log (1-D(x, z)) dxdz\end{aligned}\end{equation}
or
\begin{equation}\label{eq:our-ali-g-o2}\begin{aligned}-&\iint p(z|x)\tilde{p}(x)\log (1-D(x, z)) dxdz \\
-&\iint q(x|z) q(z)\log D(x, z) dxdz\end{aligned}\end{equation}
both of which are not like $\eqref{eq:our-ali-g}$. Our experiment shows $\eqref{eq:our-ali-g}$ has the same performance as $\eqref{eq:our-ali-g-o1}$ and $\eqref{eq:our-ali-g-o2}$. That means treating adversarial networks as a a min-max problem is not the only one approach. Variational inference may give us some new insight sometimes.

\section{Conclusion}

Our results prove that variational inference is a general framwork to derivate and explain many generative models, including VAE and GAN. We also discribe how variational inference do that by introducing a new interpretation of variational inference. This interpretation is powerful, which can lead to VAE and EM algorithm in serveral lines and deduce GAN in clearly.

An related work is \cite{Hu2018On}, which also attemps to link VAE and GAN with variational inference. However, their processing is not clear enough. we made up for this deficiency, trying to give an simpler view on GAN under variational inference.

It seems the potential of variational inference is waitting to be mined.

\bibliographystyle{aaai}
\bibliography{biblio.bib}

\end{document}